\documentclass{article}

\PassOptionsToPackage{numbers,compress}{natbib}
\usepackage[preprint]{neurips_2026}

\usepackage[utf8]{inputenc}
\usepackage[T1]{fontenc}
\usepackage{hyperref}
\usepackage{url}
\usepackage{booktabs}
\usepackage{amsfonts}
\usepackage{amsmath}
\usepackage{amssymb}
\newcommand{\best}[1]{\textcolor{red}{\textbf{#1}}}
\newcommand{\sbest}[1]{\textcolor{green!55!black}{\textbf{#1}}}
\usepackage{nicefrac}
\usepackage{float}
\usepackage{microtype}
\usepackage{xcolor}
\usepackage{graphicx}
\usepackage{bm}
\usepackage{multirow}
\usepackage[font=small,labelfont=bf,labelsep=period,singlelinecheck=false,justification=justified]{caption}
\usepackage[section]{placeins}  
\usepackage{titlesec}

\graphicspath{{figures/}{f:/3DGS_data/NIPS2026/figures/}}

\title{PD-4DGS: Progressive Decomposition of 4D Gaussian Splatting for Bandwidth-Adaptive Dynamic Scene Streaming}

\author{%
  \begin{tabular}{cccc}
    Jiachen Li & Guangzhi Han & Jin Wan & Delong Han \\[2pt]
    Yuan Gao & Min Li & Mingle Zhou & Gang Li\thanks{Corresponding author.} \\
  \end{tabular}\\[4pt]
  Qilu University of Technology
}

\begin{document}

\maketitle

\begin{center}
\includegraphics[width=\linewidth]{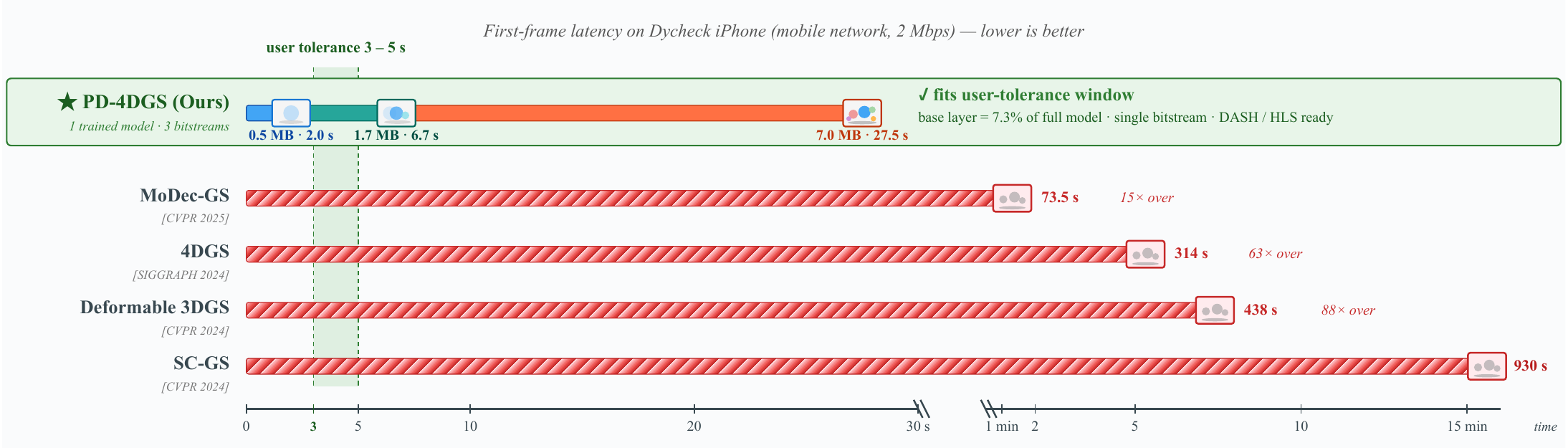}
\captionof{figure}{\textbf{PD-4DGS breaks the interactive deployment barrier of 4DGS.} First-frame latency on Dycheck iPhone over a 2~Mbps link: existing methods need 73--930~s---far beyond the 3--5~s user-tolerance window---while PD-4DGS streams a 0.44~MB static base layer in 1.7~s and progressively upgrades to global motion (1.62~MB at 6.5~s) and full HD detail (6.90~MB at 27.6~s) from a single trained model natively compatible with DASH/HLS.}
\label{fig:teaser}
\end{center}
\begin{abstract}
4D Gaussian Splatting (4DGS) enables high-quality dynamic novel view synthesis, yet current models remain monolithic bitstreams that clients must download \emph{in full} before any frame can be rendered, causing black-screen waits of tens to hundreds of seconds on mobile bandwidth and leaving 4DGS incompatible with modern adaptive-bitrate delivery. Progressive 3DGS compression alleviates this for static scenes, but it acts only on spatial anchors and cannot partition the temporal deformation networks that dominate dynamic-scene size. We present PD-4DGS, the first framework for progressive compression and on-demand transmission of 4DGS. Hierarchical Deformation Decomposition (HDD) externalises the coarse-to-fine motion hierarchy already latent in 4DGS into three independently transmittable layers---a static scaffold, a global deformation, and a local refinement---so that any prefix of the bitstream is already renderable, turning a single training run into a scalable, DASH/HLS-compatible bitstream. A Gaussian-entropy attribute rate--distortion loss together with a temporal mask consistency regulariser shrink the base layer while suppressing low-bitrate flicker; a capacity-weighted rollout schedule, gated online by a \emph{learnt} activation rate $\rho$, then prevents deformation-network under-training without any per-scene hyperparameter. On the Dycheck iPhone benchmark, PD-4DGS cuts the streamed bitstream by $>$60\% at matched rendering fidelity and reduces first-frame latency from 73--930~s to $\sim$1.7~s on a 2~Mbps link, uniquely enabling true on-demand progressive streaming for 4DGS.
\end{abstract}

\section{Introduction}

3D Gaussian Splatting (3DGS)~\citep{kerbl2023threed} delivers photorealistic real-time novel view synthesis and is rapidly extending to dynamic scenes (4DGS)~\citep{yang2023realtime4dgs,duan2024fourdrotor}. Mainstream dynamic methods couple a canonical 3DGS with a deformation network modelled via an MLP~\citep{huang2024scgs,yang2024deformable,liang2024gaufre,park2025splinegs} or feature grids~\citep{wu2024fourdgs,duisterhof2023mdsplatting,lin2024gaussianflow}, with recent work~\citep{kratimenos2025dynmf,gao2024relaygs} decomposing motion into global and local components; these methods deliver state-of-the-art fidelity but require massive storage for millions of Gaussians and their deformation fields~\citep{li2024spacetime,cho2024fourdscaffold,lee2025omg4}.

Deploying such models in streaming applications, however, is severely impeded by an ``all-or-nothing'' bottleneck: a typical two-stage 4DGS packs $\sim$15~MB of static anchors, $\sim$1.5~MB global deformation, and $\sim$8~MB local deformation into a single $\sim$25~MB bitstream, incurring $\sim$100~s of black-screen wait on a 2~Mbps mobile link. Existing 3DGS compression~\citep{fan2024lightgaussian,niedermayr2024compressed,morgenstern2023compact} and concurrent 4DGS-specific compressors~\citep{zhang2025mega,liu2025light4gs,ho2025ted4dgs,lee2025tsa4dgs} are \emph{single-rate} (retraining required to switch bitrate) and act only on spatial or per-frame primitives, so they cannot address the temporal deformation networks that dominate dynamic models. Recent progressive 3DGS compression~\citep{chen2024hac,disario2025gode,shi2024lapisgs,zoomers2024progs,huang2024hierarchical} and Gaussian-streaming systems~\citep{sun2025lts,kim2025vega,gong2025adaptive3dgs} provide multi-rate bitstreams but define progressivity purely along spatial dimensions or as an external scalability layer atop an already-trained model; a direct port to dynamic scenes fails because static masks cannot meaningfully partition a temporal deformation network.

We present PD-4DGS, the first framework for progressive compression and on-demand transmission of 4DGS. Our key observation is that high-quality 4DGS~\citep{katsumata2024compact,katsumata2023efficient} already adopts a coarse-to-fine physical hierarchy---a global stage for large-scale motion and a local stage for high-frequency detail---that we \emph{externalise} into three independently transmittable bitstream layers (Hierarchical Deformation Decomposition, HDD): a Static Scaffold, a Global Deformation layer, and a Local Refinement layer, so that any prefix is already renderable as a static snapshot, a coarsely animated scene, or full-fidelity dynamics. Externalising the hierarchy, however, surfaces two challenges that static progressive compression never has to face. First, all three layers must share a single anchor set, yet the base layer is required to be MB-scale for second-scale first-frame latency while deeper layers demand much richer per-anchor attributes; naively pruning attributes for the base layer destroys the fidelity of the deeper layers, and pruning conservatively inflates the base layer beyond its latency budget. Second, dynamic rendering threads attribute-level masks through a temporal deformation network: tiny mask differences across frames are amplified into visible inter-frame jitter, so a static-scene quantisation/masking scheme that looks adequate per-frame becomes a flicker source as soon as the deformation network is rolled out over time. We address these two issues with an Attribute-Level Rate--Distortion Optimisation (R-DO) and a Temporal Mask Consistency (TMC) regulariser, deferring algorithmic details to Sec.~\ref{sec:method}.

Uniform layer sampling would systematically under-train the higher-capacity deformation networks, collapsing progressivity into a pseudo-progressive regime. We therefore propose Capacity-Weighted Progressive Rollout Training, which samples layers non-uniformly and drives the sampling distribution online via a \emph{learnt} mask-activation rate $\rho = \mathbb{E}[m_{\text{top}} - m_{\text{base}}]$; the adaptive distribution $\pi(\rho) = (1{-}\rho)\,\pi_{\text{uniform}} + \rho\,\pi_{\text{aggressive}}$ reallocates budget across levels according to each scene's deformation complexity, without any per-scene hyperparameter.

Extensive experiments on the Dycheck iPhone monocular benchmark demonstrate that PD-4DGS achieves a streamed bitstream reduction of over 60\% relative to the strongest dynamic baseline at matched rendering fidelity, delivers a renderable static first frame within two seconds on a 2~Mbps mobile link, and restores strict layer-wise monotonicity of progressive decoding on every scene. The key contributions of this work are summarised as follows:

\begin{itemize}
\item We present PD-4DGS, the first progressive 4DGS framework, whose Hierarchical Deformation Decomposition (HDD) factorises the rendering network into static, global, and local layers, achieving \emph{network-level} rather than spatial-only progressivity.
\item We design an attribute-level rate--distortion loss with a temporal mask consistency regulariser that compresses anchors via a differentiable Gaussian entropy model and suppresses low-bitrate flicker.
\item We propose capacity-weighted rollout training driven by a $\rho$-adaptive sampling distribution $\pi(\rho)$ that resolves deformation-network under-training \emph{without any per-scene hyperparameter}, cutting the streamed volume by $>$60\% on Dycheck iPhone at matched fidelity.
\end{itemize}

\section{Related Work}

\paragraph{4DGS.}
Dynamic extensions of 3DGS~\citep{kerbl2023threed} either treat time as an additional Gaussian dimension~\citep{duan2024fourdrotor,liu2024modgs} or couple a canonical 3DGS with a deformation network modelled implicitly via an MLP~\citep{huang2024scgs,yang2024deformable,liang2024gaufre,park2025splinegs} or explicitly via feature grids~\citep{wu2024fourdgs,duisterhof2023mdsplatting,lin2024gaussianflow}. To handle complex real-world motion, recent methods~\citep{kratimenos2025dynmf,gao2024relaygs} factor the motion into global and local components---a representative example being the two-stage ``global anchor + local Gaussian'' pipeline of MoDec-GS~\citep{katsumata2024compact}. This hierarchical structure, designed for reconstruction quality, is the observation our work externalises for transmission. Despite their fidelity, these models are monolithic and storage-heavy~\citep{li2024spacetime,cho2024fourdscaffold,lee2025omg4}, which rules out bandwidth-adaptive streaming.

\paragraph{3DGS compression and progressive streaming.}
Static-3DGS compression combines pruning~\citep{fan2024lightgaussian,niedermayr2024compressed}, vector quantisation~\citep{navaneet2024compact3d,lee2024compact}, and context modelling~\citep{lu2024scaffold,morgenstern2023compact,chen2024hac,wang2024contextgs,zhan2025cat3dgs,liu2024hemgs,girish2024eagles,lee2024compactstaticdynamic}; concurrent 4DGS-specific compressors~\citep{zhang2025mega,liu2025light4gs,ho2025ted4dgs,lee2025tsa4dgs} extend the same rate--distortion machinery to dynamic primitives. All of them remain \emph{single-rate}, operate only on spatial/per-frame anchors, and offer no mechanism for the deformation networks that often contribute $>$40\% of a dynamic model. Progressive variants~\citep{cheng2024gaussianpro,zoomers2024progs,shi2024lapisgs,disario2025gode,huang2024hierarchical,jeon2023context} and dynamic streaming solutions~\citep{sun2024threedgstream,liu2024swings,sun2025lts,kim2025vega,gong2025adaptive3dgs} turn a single training run into a multi-level scalable bitstream, but progressivity is defined purely along spatial dimensions or layered \emph{externally} on top of an already-trained model, never performing a progressive decomposition of the representation itself. PD-4DGS fills this gap by decomposing the \emph{rendering pipeline} rather than only its spatial anchors.

\paragraph{Progressive and capacity-aware training.}
Hierarchical networks benefit from curriculum learning~\citep{xiangli2022bungeenerf,cheng2024gaussianpro}, but progressive 3DGS frameworks~\citep{chen2024hac,shi2024lapisgs} all sample layers uniformly at training time. When deformation networks are several times larger than the MLP head, uniform sampling systematically under-trains the high-capacity branch---an instance of the capacity--sampling mismatch reported in MoE~\citep{girish2024eagles}. Our capacity-weighted rollout (Sec.~\ref{sec:rollout}) closes this gap by sampling layers non-uniformly with frequencies commensurate with their parameter count.

\paragraph{Positioning.}
In summary, PD-4DGS differs from prior work along three axes: unlike monolithic 4DGS~\citep{huang2024scgs,yang2024deformable,wu2024fourdgs,katsumata2024compact,park2025splinegs,lee2025omg4} it emits a \emph{scalable}, ABR-compatible bitstream; unlike single-rate or progressive 3DGS/4DGS compression~\citep{fan2024lightgaussian,niedermayr2024compressed,chen2024hac,shi2024lapisgs,disario2025gode,zhang2025mega,liu2025light4gs,ho2025ted4dgs,lee2025tsa4dgs} and externally layered streaming systems~\citep{sun2025lts,kim2025vega,gong2025adaptive3dgs}, it operates on the \emph{deformation network} and adapts the per-layer training budget online to each scene's dynamic complexity, making the representation \emph{intrinsically} scalable.

\section{Method}
\label{sec:method}

\paragraph{Notation.}
The canonical anchor set is $A = \{(x_v, f_v, s_v, o_v)\}_{v \in V}$ (position, feature, scale, offset); each anchor carries a learnable mask $m_i \in [0,1]$. The dynamic pipeline chains a global deformation network $\mathcal{F}_{\text{global}}$, a local deformation network $\mathcal{F}_{\text{local}}$, and a rasteriser $\mathcal{R}$, giving the full forward pass $\mathcal{R}(\text{MLP}(A + \mathcal{F}_{\text{global}}(A,t)) + \mathcal{F}_{\text{local}}(G,t))$ where $G$ denotes the neural Gaussians produced from the anchors.

\begin{figure}[!htbp]
\centering
\includegraphics[width=\linewidth]{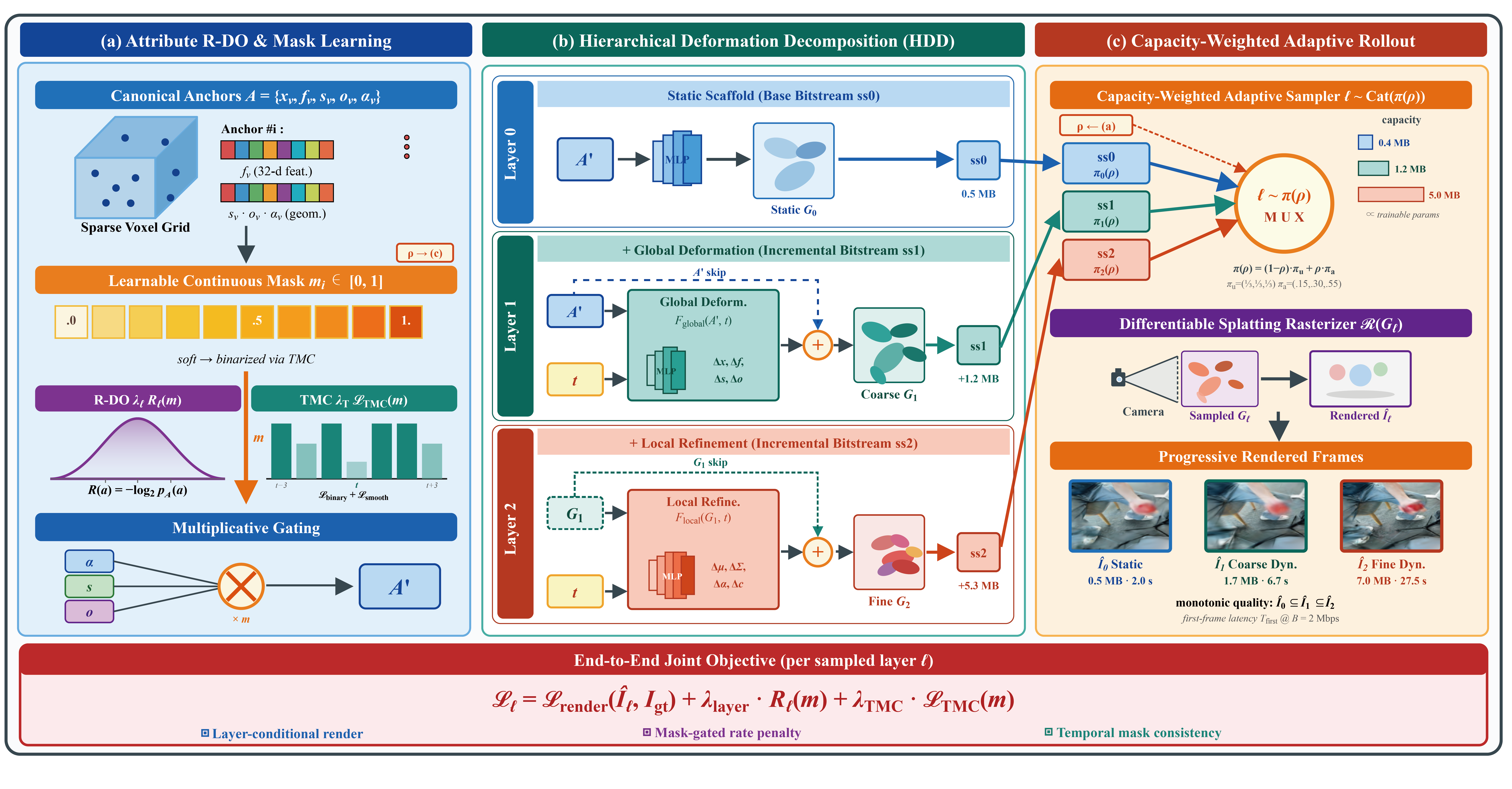}
\caption{\textbf{PD-4DGS pipeline.} (a)~Attribute R-DO compresses anchor attributes through a learnable mask $m_i$ and a Gaussian entropy model, while temporal mask consistency (TMC) stabilises $m_i$ over time; the gated anchor set $A'$ is the input to (b). (b)~HDD factorises the rendering network into three additive bitstreams---Static Scaffold (ss0), $+$Global Deformation (ss1), $+$Local Refinement (ss2). (c)~A capacity-weighted sampler $\ell\sim\pi(\rho)$, driven online by the mask-activation rate $\rho$, shifts training budget toward deeper layers on motion-dense scenes; the multiplexer routes the chosen bitstream to the rasteriser, producing $\hat{I}_\ell$.}
\label{fig:pipeline}
\end{figure}

PD-4DGS jointly optimises, in a single end-to-end pass, four components: HDD (Sec.~\ref{sec:hdd}), R-DO (Sec.~\ref{sec:rdo}), TMC (Sec.~\ref{sec:tmc}), and capacity-weighted rollout training (Sec.~\ref{sec:rollout}); Fig.~\ref{fig:pipeline} gives the overview.

\subsection{Hierarchical Deformation Decomposition}
\label{sec:hdd}

4DGS architectures~\citep{katsumata2024compact,katsumata2023efficient} expose their internal coarse-to-fine hierarchy only as an inductive bias and pack the result into a monolithic bitstream; HDD externalises the same hierarchy at the bitstream level as three independently transmittable layers.

\textbf{Layer~0 (Static Scaffold)} carries only the canonical anchors and the attribute-prediction MLP, so the client can render a static snapshot immediately on arrival of this layer without any temporal input. \textbf{Layer~1 (Global Deformation)} incrementally transmits $\mathcal{F}_{\text{global}}$ (a lightweight HexPlane + MLP) that predicts per-anchor displacements and feature residuals, lifting the scene into dynamic mode; because the deformation acts on \emph{sparse} anchors, Layer~1 typically adds $<$10\% of total model volume yet already captures the dominant motion. \textbf{Layer~2 (Local Refinement)} transmits $\mathcal{F}_{\text{local}}$, which predicts per-Gaussian residuals $(\Delta\mu, \Delta\Sigma, \Delta\alpha, \Delta c)$ on top of the coarse dynamic Gaussians, recovering high-frequency detail (hair, facial muscle motion) for full fidelity. Depending on which prefix $s\in\{0,1,2\}$ has been received, the renderer routes the computation graph
\begin{equation}
\text{Render}_s(t) = 
\begin{cases}
\mathcal{R}(\text{MLP}(A)), & s = 0, \\[2pt]
\mathcal{R}(\text{MLP}(A + \mathcal{F}_{\text{global}}(A, t))), & s = 1, \\[2pt]
\mathcal{R}(\text{MLP}(A + \mathcal{F}_{\text{global}}(A, t)) + \mathcal{F}_{\text{local}}(G, t)), & s = 2.
\end{cases}
\end{equation}
All three levels share the same anchors and MLP, so switching levels amounts to additively loading further weights---no re-decoding or retraining---which makes PD-4DGS natively adaptive to bandwidth fluctuations and heterogeneous-device deployment from a single bitstream.

\subsection{Attribute-Level Rate--Distortion Optimization}
\label{sec:rdo}

Layer~0 carries $\sim$60\% of total model size, so a bulky base layer defeats the second-scale first-frame promise. We cast keep/drop as a differentiable rate--distortion problem in which bit rate is an explicit loss variable, rather than relying on heuristic pruning thresholds~\citep{fan2024lightgaussian,niedermayr2024compressed}.

\paragraph{Learnable soft mask.}
For each anchor $i$ we introduce a learnable mask $m_i \in [0,1]$ that symmetrically multiplies both the Gaussian's opacity and scale, $\alpha'_i = \alpha_i m_i$ and $s'_i = s_i m_i$, in every forward pass at both training and inference time; at inference we hard-threshold $m_i > 0.01$. This dual multiplicative gating ensures an anchor simultaneously becomes transparent and volume-zero as $m_i \to 0$ and eliminates the ``soft-train/hard-prune'' distribution mismatch that plagues heuristic pruning~\citep{fan2024lightgaussian,niedermayr2024compressed}.

\paragraph{Gaussian entropy model.}
For each continuous attribute $a \in \{f_{v,i}, s_{v,i}, o_{v,i}\}$ we assume a per-batch Gaussian prior $\mathcal{N}(\mu_a, \sigma_a^2)$ estimated on the \emph{active} anchor subset, giving Shannon bit cost
\begin{equation}
R(a) = -\log_2\!\left[\Phi\!\left(\tfrac{a + Q_a/2 - \mu_a}{\sigma_a}\right) - \Phi\!\left(\tfrac{a - Q_a/2 - \mu_a}{\sigma_a}\right)\right],
\end{equation}
under a fixed quantisation step $Q_a$ and standard-Gaussian CDF $\Phi$ (clipped by $10^{-6}$). Unlike plain $\ell_1/\ell_2$ regularisation, this term \emph{directly} approximates an entropy coder's output length (LZMA at export), tightly coupling the training-time compression signal to the exported bitstream size.

\paragraph{Layer-wise rate--distortion objective.}
Combining distortion and bit cost yields
\begin{equation}
\mathcal{L}_{\text{total}} = \mathcal{L}_{\text{render}} + \lambda_{\text{layer}} \cdot \tfrac{1}{|V|} \sum_{i \in V} m_i \cdot \big(R(f_{v,i}) + R(s_{v,i}) + R(o_{v,i})\big),
\end{equation}
where $\mathcal{L}_{\text{render}}$ is an $\ell_1$+D-SSIM loss. Only active anchors ($m_i>0$) incur a bit budget, giving the optimiser an unambiguous keep/drop signal: if an anchor's distortion contribution fails to offset its rate, $m_i$ is driven toward zero. For multi-level compression we use $\lambda_{\text{layer}} \in \{0.04, 0.01, 0.00025\}$---roughly an order-of-magnitude separation---so Layer~0 is aggressively compressed while Layer~2 preserves almost all anchors for maximum fidelity.

\subsection{Temporal Mask Consistency}
\label{sec:tmc}

R-DO sparsification induces high-frequency mask oscillation along the temporal axis---flicker and ghost trails that $\ell_1$ sparsity cannot capture. TMC stabilises the activation pattern with two complementary regularisers.

\paragraph{Binary entropy penalty.}
A Bernoulli binary-entropy penalty drives mask values deterministically toward $\{0, 1\}$,
\begin{equation}
\mathcal{L}_{\text{binary}} = -\tfrac{1}{|V|} \sum_{i \in V} \big(m_i \log_2 m_i + (1 - m_i) \log_2 (1 - m_i)\big),
\end{equation}
eliminating semi-transparent ghost trails and minimising the entropy-coder code length.

\paragraph{Spatial smoothness constraint.}
Exploiting the prior that \emph{spatially adjacent anchors share activation states}, we add
\begin{equation}
\mathcal{L}_{\text{smooth}} = \tfrac{1}{|S|} \sum_{(i,j) \in S} w_{ij} |m_i - m_j|, \quad w_{ij} = \exp(-\|x_i - x_j\|/\tau),
\end{equation}
on randomly sampled anchor pairs $S$, which implicitly enforces cross-frame consistency without directly modelling the temporal axis. The combined loss $\mathcal{L}_{\text{TMC}} = \mathcal{L}_{\text{binary}} + \mathcal{L}_{\text{smooth}}$ (weight $\lambda_{\text{temporal}}$) keeps Layer~0's anchor selection coherent over time; the ablation in Sec.~\ref{sec:ablation} shows the two terms are non-interchangeable.

\subsection{Capacity-Weighted Progressive Rollout Training}
\label{sec:rollout}

Architecture and loss alone do not deliver layer-wise monotonic refinement; the training-time layer-sampling strategy is equally critical.

\paragraph{Layer collapse and the limits of fixed priors.}
Let $\ell \in \{0,1,2\}$ index the Base, +GAD, and +GAD+LGD forward modes and suppose $\ell \sim \text{Categorical}(\pi)$. Under the natural choice $\pi = (1/3,1/3,1/3)$, the higher-capacity GAD/LGD networks (2--6$\times$ the base-layer parameters) are back-propagated through only a third of the time: on the Dycheck spin scene we observe $\ell_0/\ell_1/\ell_2$ PSNRs of 14.61/14.54/14.54, i.e.\ deeper levels fail to outperform the base and progressivity collapses. A fixed capacity-weighted prior such as $\pi_{\text{aggr}} = (0.15, 0.30, 0.55)$ fixes the dynamic scenes ($+0.13$~dB on spin, $+0.22$~dB on paper-windmill, matching MoE-style balancing~\citep{girish2024eagles}) but \emph{regresses} by 0.57~dB on the static/dynamic-mixed block scene because the low $\pi_0$ under-trains the sparse scaffold: no single $\pi$ is optimal for all scenes.

\paragraph{A mask-driven adaptive sampling distribution $\pi(\rho)$.}
The progressive masks $m_i^{(\ell)}$ already encode, during training, each scene's demand for deeper-deformation budget: anchors activated only at the top level imply rich local motion, while anchors active at the base layer imply static dominance. We therefore define the inter-level mask activation rate $\rho = \mathbb{E}_{i \in V}[m_i^{(\ell=2)} - m_i^{(\ell=0)}] \in [0,1]$, a fully learnt statistic requiring no external annotation, and use it as the interpolation coefficient between two fixed end distributions:
\begin{equation}
\pi(\rho) \;=\; (1 - \rho)\,\pi_{\text{uniform}} \;+\; \rho\,\pi_{\text{aggr}},
\quad \pi_{\text{uniform}} = (\tfrac{1}{3},\tfrac{1}{3},\tfrac{1}{3}),\; \pi_{\text{aggr}} = (0.15, 0.30, 0.55).
\end{equation}
As $\rho \to 0$ (static) $\pi \to \pi_{\text{uniform}}$ so the base layer is trained well; as $\rho \to 1$ (highly dynamic) $\pi \to \pi_{\text{aggr}}$ and the deeper networks receive enriched signal, with smooth interpolation in between.

To smooth the early-training $m^{(\ell)}$ noise, we sample $\hat\rho_t$ every $K{=}200$ iterations and drive $\pi$ from the EMA $\rho_t^{\text{ema}} = (1{-}\alpha)\rho_{t-1}^{\text{ema}} + \alpha\hat\rho_t$ ($\alpha{=}0.05$) after a warm-up $T_{\text{warm}}{=}2000$ during which $\pi_{\text{uniform}}$ is used. The constants $\{K, \alpha, T_{\text{warm}}, \pi_{\text{uniform}}, \pi_{\text{aggr}}\}$ are identical across all scenes---every inter-scene difference is absorbed by the trained $\rho$.

\paragraph{Training objective.}
Given the sampled $\ell$, the per-iteration forward $F_\ell(A,t)$ uses exactly the same computation graph as $\text{Render}_\ell(t)$ in Sec.~\ref{sec:hdd}; this symmetry between training and inference, together with symmetric mask gating, removes the distribution shift that plagues frameworks~\citep{chen2024hac,shi2024lapisgs} running the full $\ell{=}2$ path at train time and switching levels only at inference. The loss for the sampled level is
\begin{equation}
\mathcal{L}_\ell = \mathcal{L}_{\text{render}}(F_\ell(A,t)) + \lambda_\ell R_\ell(m) + \lambda_{\text{temporal}} \mathcal{L}_{\text{TMC}}(m),
\end{equation}
with $R_\ell(m) = \tfrac{1}{|V|} \sum_i m_i^{(\ell)} (R(f_{v,i}) + R(s_{v,i}) + R(o_{v,i}))$, and the global expected loss $\mathbb{E}_{\ell \sim \pi(\rho)}[\mathcal{L}_\ell]$ implements scene-aware budget allocation through $\pi(\rho)$: because sampling adapts to scene complexity, all three layers converge at comparable rates and the monotonicity $\text{PSNR}(\ell_0) \le \text{PSNR}(\ell_1) \le \text{PSNR}(\ell_2)$ is restored on all 7 Dycheck-iPhone scenes (Sec.~\ref{sec:exp}).

\section{Experiments}
\label{sec:exp}

\subsection{Experimental Setup}

\paragraph{Implementation and metrics.}
PD-4DGS is implemented in PyTorch on a single RTX 4070 Super (12~GB VRAM); the base architecture~\citep{kratimenos2025dynmf,lu2024scaffold} uses $L{=}3$ progressive levels and 30k iterations per scene. The first 10k iterations use only $\mathcal{L}_{\text{render}} = \mathcal{L}_1 + \lambda_{\text{D-SSIM}}\mathcal{L}_{\text{D-SSIM}}$; from $T_{\text{progressive}}{=}10{,}000$ we enable R-DO and TMC with $\lambda_{\text{temporal}}{=}0.01$ and export each layer with LZMA. We evaluate on the Dycheck iPhone monocular benchmark~\citep{gao2022monocular} and report the official masked variants mPSNR/mSSIM/mLPIPS inside the co-visible region, the streamed bitstream volume (\emph{Stream}), and the frame-to-frame LPIPS difference \emph{t-LPIPS} for temporal stability.

\subsection{Comparative Experiments}

Baselines: SC-GS~\citep{huang2024scgs} (sparse-control), Deformable 3DGS~\citep{yang2024deformable}, 4DGS~\citep{wu2024fourdgs}, and our backbone MoDec-GS~\citep{katsumata2024compact}; per-column per-metric, the best is \best{red} and the second-best \sbest{green}.

\begin{table}[H]
\caption{Quantitative comparison on Dycheck iPhone~\citep{gao2022monocular}. Each cell shows mPSNR / mSSIM / mLPIPS / Stream(MB). Values in \best{red} mark the best result and values in \sbest{green} the second-best result for each per-column per-metric entry.}
\label{tab:main}
\centering
\scriptsize
\setlength{\tabcolsep}{3pt}
\resizebox{\linewidth}{!}{%
\begin{tabular}{l cccc}
\toprule
Method & Apple & Block & Paper-windmill & Space-out \\
\midrule
SC-GS              & 14.96 / 0.692 / 0.508 / 173.3 & 13.98 / 0.548 / 0.483 / 115.7 & 14.87 / \best{0.221} / 0.432 / 446.3 & \best{14.79} / 0.511 / \sbest{0.440} / 114.2 \\
Deformable 3DGS    & 15.61 / \sbest{0.696} / \best{0.367} / 87.71 & 14.87 / 0.559 / \best{0.390} / 118.9 & 14.89 / 0.213 / \best{0.341} / 160.2 & 14.59 / 0.510 / 0.450 / 42.01 \\
4DGS               & 15.41 / 0.691 / 0.524 / 61.52 & 13.89 / 0.550 / 0.539 / 63.52 & 14.44 / 0.201 / 0.445 / 123.9 & 14.29 / \sbest{0.515} / 0.473 / 52.02 \\
MoDec-GS           & \sbest{16.48} / \best{0.699} / \sbest{0.402} / \sbest{23.78} & \best{15.57} / \best{0.590} / \sbest{0.478} / \sbest{13.65} & \sbest{14.92} / \sbest{0.220} / \sbest{0.377} / \sbest{17.08} & \sbest{14.65} / \best{0.522} / 0.467 / \sbest{18.24} \\
PD-4DGS (Ours) & \best{17.01} / 0.695 / 0.500 / \best{6.58} & \sbest{15.02} / \sbest{0.587} / 0.553 / \best{6.65} & \best{14.98} / 0.217 / 0.389 / \best{6.98} & 14.61 / \best{0.522} / \best{0.426} / \best{7.23} \\
\bottomrule
\end{tabular}}

\vspace{0.7em}

\resizebox{\linewidth}{!}{%
\begin{tabular}{l cccc}
\toprule
Method & Spin & Teddy & Wheel & Average \\
\midrule
SC-GS              & 14.32 / 0.407 / 0.445 / 219.1 & 12.51 / \sbest{0.516} / \best{0.562} / 318.7 & 11.90 / 0.354 / 0.484 / 239.2 & 13.90 / \sbest{0.464} / 0.479 / 232.4 \\
Deformable 3DGS    & 13.10 / 0.392 / 0.490 / 133.9 & 11.20 / 0.508 / \sbest{0.573} / 117.1 & 11.79 / 0.345 / \best{0.394} / 106.1 & 13.72 / 0.461 / \best{0.430} / 109.4 \\
4DGS               & 14.89 / 0.413 / 0.441 / 71.80 & 12.31 / 0.509 / 0.605 / 80.44 & 10.83 / 0.339 / 0.538 / 96.50 & 13.72 / 0.460 / 0.509 / 78.54 \\
MoDec-GS           & \best{15.53} / \sbest{0.433} / \best{0.366} / \sbest{26.84} & \sbest{12.56} / \best{0.521} / 0.598 / \sbest{12.28} & \sbest{12.44} / \sbest{0.374} / \sbest{0.413} / \sbest{16.68} & \best{14.60} / \best{0.480} / \sbest{0.443} / \sbest{18.37} \\
PD-4DGS (Ours) & \sbest{14.97} / \best{0.450} / \sbest{0.387} / \best{7.32} & \best{12.66} / 0.515 / 0.648 / \best{6.76} & \best{12.50} / \best{0.375} / 0.464 / \best{6.64} & \sbest{14.54} / \best{0.480} / 0.481 / \best{6.88} \\
\bottomrule
\end{tabular}}
\end{table}
\begin{figure}[H]
\centering
\includegraphics[width=\linewidth]{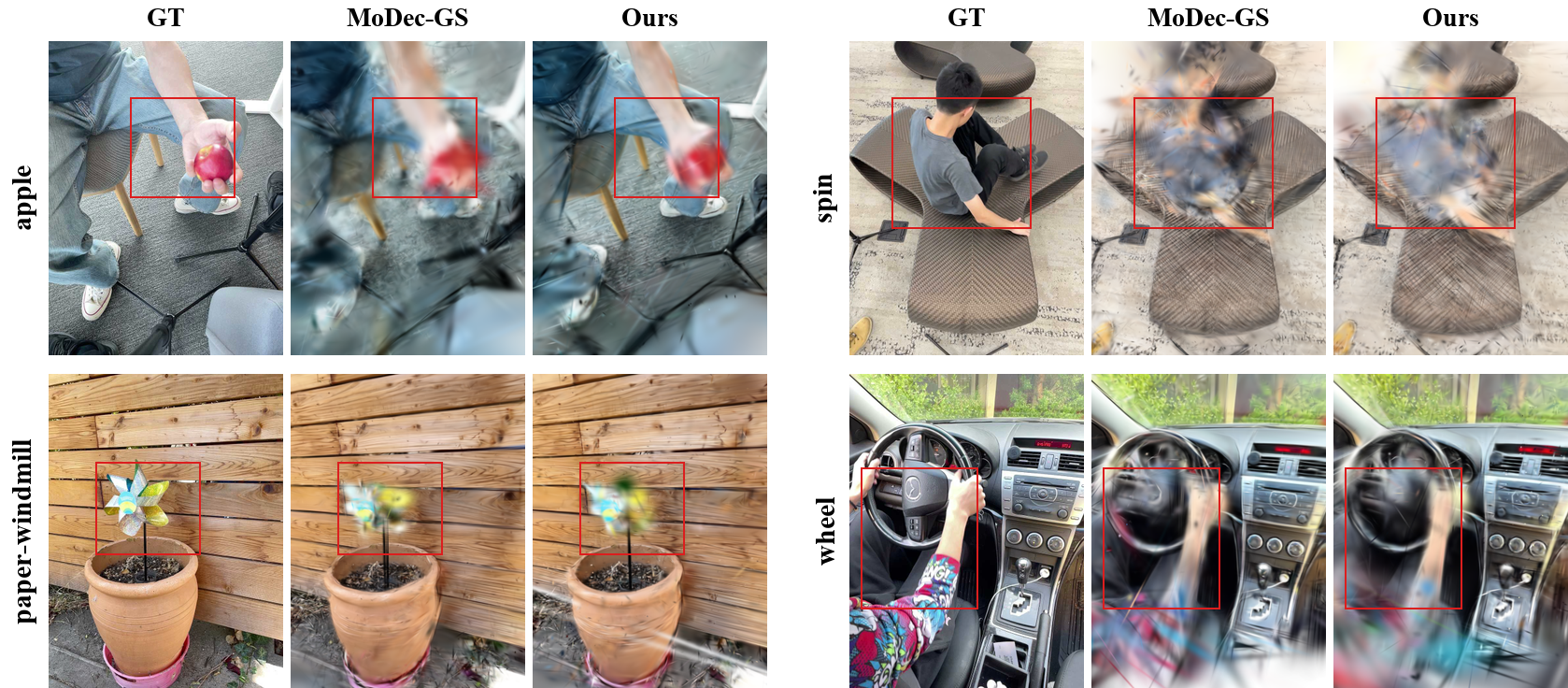}
\caption{\textbf{Qualitative comparison on Dycheck iPhone.} Four scenes (\textit{apple}, \textit{spin}, \textit{paper-windmill}, \textit{wheel}; vertical labels) in a $2{\times}2$ grid; within each block: GT, MoDec-GS (18.37~MB), Ours (6.88~MB). \textcolor{red}{\textbf{Red boxes}} mark the region of interest, identical across columns. At $\sim$62\% smaller bitstream, Ours matches MoDec-GS on static-leaning/high-frequency-edge cases and inherits the same residual motion artefacts on the hardest fast-rotation case (\textit{spin}).}
\label{fig:qualitative}
\end{figure}

Table~\ref{tab:main} (cells: mPSNR~/~mSSIM~/~mLPIPS~/~Stream~MB) shows PD-4DGS averaging only 6.88~MB of streamed bitstream---a 62.6\% reduction relative to MoDec-GS (18.37~MB) and $\sim$$11\times$/$34\times$ smaller than 4DGS (78.54~MB) and SC-GS (232.4~MB)---at matched rendering quality: average mPSNR 14.54~dB is within 0.06~dB of MoDec-GS (14.60) while surpassing it on Apple, Paper-windmill, Teddy, and Wheel, and average mSSIM (0.480) matches MoDec-GS exactly. The $\rho$-adaptive rollout of Sec.~\ref{sec:rollout} recovers $+0.20$~dB on the static/dynamic-mixed Block scene over a fixed capacity-weighted prior (Sec.~\ref{sec:ablation}), validating scene-heterogeneity handling. Average mLPIPS (0.481 vs.\ 0.443) is slightly higher, mainly from the 60\% volume reduction trimming high-frequency detail on Block/Teddy---a trade-off controllable via $\lambda_0$. Fig.~\ref{fig:qualitative} corroborates: at 6.88~MB our renderings preserve the textured object in \textit{apple}, windmill blades and wood-fence high-frequency edges in \textit{paper-windmill}, and the steering-wheel rim in \textit{wheel} on par with MoDec-GS at 18.37~MB, and both methods share the same residual motion blur on the hardest case (\textit{spin})---the bitstream reduction introduces no new failure mode.

\FloatBarrier
\subsection{Ablation Studies}
\label{sec:ablation}

We ablate the sampling strategy of rollout, R-DO, and TMC. The sampling ablation (\ref{sec:abl-sampling}) uses the deformation-dense subset (block, spin, paper-windmill); the component (\ref{sec:abl-comp}) and monotonicity (\ref{sec:abl-mono}) studies use all 7 Dycheck scenes. All variants share the network, schedule, and compression pipeline of the main model.

\subsubsection{Effectiveness of $\rho$-Adaptive Sampling}
\label{sec:abl-sampling}

We compare $\pi(\rho)$ (Sec.~\ref{sec:rollout}) against two naive baselines: Uniform $\pi{=}(1/3,1/3,1/3)$ and CW-Fixed $\pi_{\text{aggr}}{=}(0.15, 0.30, 0.55)$.

\begin{table}[!ht]
\caption{Sampling-strategy ablation. Cells show mPSNR / Stream(MB).}
\label{tab:abl-sampling}
\centering
\footnotesize
\begin{tabular}{l ccc cc}
\toprule
Strategy & Block & Spin & Paper-windmill & Avg.\ mPSNR & Avg.\ Stream (MB) \\
\midrule
Uniform $\pi = (\tfrac{1}{3}, \tfrac{1}{3}, \tfrac{1}{3})$       & 15.39 / 6.69 & 14.80 / 7.38 & 14.84 / 7.01 & 15.01 & 7.03 \\
CW-Fixed $\pi = (.15, .30, .55)$ & 14.82 / 6.91 & 14.93 / 7.59 & 15.06 / 7.05 & 14.94 & 7.18 \\
\textbf{CW-Adaptive (Ours)}      & \textbf{15.02} / \textbf{6.65} & \textbf{14.97} / \textbf{7.32} & 14.98 / \textbf{6.98} & \textbf{14.99} & \textbf{6.98} \\
\bottomrule
\end{tabular}
\end{table}

CW-Fixed gains $+0.13$~dB and $+0.22$~dB on dynamic scenes (spin, paper-windmill) but \emph{regresses} $-0.57$~dB on the static/dynamic-mixed block scene and inflates its model by 2.2\%---no fixed $\pi$ is simultaneously optimal. CW-Adaptive interpolates between the two priors online via $\rho$, recovers $+0.20$~dB on block at the smallest size (6.65~MB), keeps essentially all of the dynamic-scene gains, and attains the best average mPSNR (14.99~dB) and smallest average stream (6.98~MB), all \emph{without any scene-specific hyperparameter}.

\subsubsection{R-DO and TMC Component Ablation}
\label{sec:abl-comp}

All variants retain HDD and CW-Adaptive; components are removed one at a time. Table~\ref{tab:abl-comp} reports the 7-scene average.

\begin{table}[!ht]
\caption{Component ablation on R-DO and TMC (Dycheck iPhone, average over 7 scenes; \textsuperscript{\dag} denotes a 6-scene average, see footnote). All variants share HDD and CW-Adaptive rollout; only the listed component is removed.}
\label{tab:abl-comp}
\centering
\footnotesize
\begin{tabular}{l ccccc}
\toprule
Variant & mPSNR & mSSIM & mLPIPS & Stream (MB) & t-LPIPS \\
\midrule
w/o R-DO                                                & 14.48 & 0.480 & \textbf{0.475} & 7.49 & 0.0149 \\
w/o $\mathcal{L}_{\text{binary}}$ (smoothness only)    & 14.48 & 0.480 & 0.480 & 6.88 & 0.0144 \\
w/o $\mathcal{L}_{\text{smooth}}$ (binary only)        & \textbf{14.58} & 0.481 & 0.482 & 6.89 & 0.0139 \\
w/o TMC ($\mathcal{L}_{\text{TMC}}$ removed)\textsuperscript{\dag}          & 14.52 & \textbf{0.488} & 0.500 & \textbf{6.81} & \textbf{0.0137} \\
\textbf{Full (Ours)}                                    & 14.54 & 0.480 & 0.478 & 6.90 & 0.0143 \\
\bottomrule
\end{tabular}
\end{table}

\textsuperscript{\dag}\,The \emph{w/o TMC} entry is averaged over 6 scenes because the Spin run did not complete its masked evaluation; the four ablation rows are otherwise comparable.

Three complementary roles emerge. \textbf{(i)} R-DO is the dominant compression force: disabling it inflates the stream from 6.90 to 7.49~MB ($+8.6\%$) for only $-0.06$~dB mPSNR. \textbf{(ii)} Within TMC, $\mathcal{L}_{\text{binary}}$ and $\mathcal{L}_{\text{smooth}}$ are non-interchangeable---dropping the former lets fractional masks oscillate ($\text{t-LPIPS}\,0.0143{\to}0.0144$), dropping the latter yields binary but isolated flicker ($+0.004$ mLPIPS). \textbf{(iii)} Removing TMC entirely lowers t-LPIPS but produces visible ghost trails and the largest mLPIPS regression ($+0.022$), showing that t-LPIPS captures temporal smoothness, not perceptual fidelity. The full model attains the best balance.

\subsubsection{Monotonicity Verification of Progressive Levels}
\label{sec:abl-mono}

Table~\ref{tab:abl-mono} evaluates mPSNR, mSSIM, t-LPIPS, and cumulative size at $\ell \in \{0,1,2\}$ with only the prefix-$\ell$ deformation networks activated, faithfully simulating layered decoding.

\begin{table}[!ht]
\caption{Monotonicity of progressive levels (Dycheck iPhone, average over all 7 scenes; same training run as Table~\ref{tab:main}).}
\label{tab:abl-mono}
\centering
\footnotesize
\begin{tabular}{l l ccc c}
\toprule
Level $\ell$ & Received component & Avg.\ mPSNR & Avg.\ mSSIM & Avg.\ t-LPIPS & Cumulative Stream (MB) \\
\midrule
$\ell = 0$ & Static Scaffold (Base)   & 14.122 & \textbf{0.354} & \textbf{0.0008} & \textbf{0.436} \\
$\ell = 1$ & + Global Deformation     & 14.175 & 0.352 & 0.0019 & 1.623 \\
$\ell = 2$ & + Local Refinement       & \textbf{14.370} & 0.339 & 0.0143 & 6.898 \\
\bottomrule
\end{tabular}
\end{table}

Average mPSNR rises strictly with the level ($14.12{\to}14.18{\to}14.37$~dB), confirming HDD's design intent---every additional deformation layer brings a quantifiable quality gain. The base layer alone is only $S_0 \approx 0.44$~MB ($\sim$6.3\% of the full model), $\ell{=}1$ reaches 1.62~MB for global motion, and $\ell{=}2$ reaches 6.90~MB for full fidelity; see Sec.~\ref{sec:streaming} for the resulting latency. The mild increase of t-LPIPS with level reflects the progressively richer dynamic content, not a quality drop: it is a well-known artefact of progressive compression whereby a static baseline appears temporally smoother than any dynamic refinement built on top of it. Critically, the monotonic ordering $\text{PSNR}(\ell_0) \le \text{PSNR}(\ell_1) \le \text{PSNR}(\ell_2)$ holds on every one of the 7 Dycheck scenes after CW-Adaptive training, even on Block where the fixed prior collapses the deeper layers---verifying that the $\rho$-adaptive rollout delivers \emph{per-scene} progressive guarantees that no fixed sampling policy can simultaneously offer.

\subsection{Progressive Streaming Analysis}
\label{sec:streaming}

We study the on-demand-loading benefit of HDD. Let $S_{\text{first}}$ be the minimum downloadable volume and $B$ (Mbps) the bandwidth; the first-frame latency lower bound is $T_{\text{first}} = 8 S_{\text{first}}/B$~(s). For monolithic baselines $S_{\text{first}}$ is the full model; for PD-4DGS Layer~0 alone already renders a static snapshot, so $S_{\text{first}}{=}S_0$. Table~\ref{tab:streaming} reports $T_{\text{first}}$ at three representative bandwidths.

\begin{table}[!ht]
\caption{First-frame latency on Dycheck iPhone.}
\label{tab:streaming}
\centering
\footnotesize
\begin{tabular}{l c ccc}
\toprule
Method & $S_{\text{first}}$ (MB) & $B = 2$ Mbps & $B = 10$ Mbps & $B = 50$ Mbps \\
\midrule
SC-GS                       & 232.4 & 929.6 s & 185.9 s & 37.2 s \\
Deformable 3DGS             & 109.4 & 437.6 s & 87.5 s  & 17.5 s \\
4DGS                        & 78.5  & 314.0 s & 62.8 s  & 12.6 s \\
MoDec-GS                    & 18.37 & 73.5 s  & 14.7 s  & 2.94 s \\
\textbf{PD-4DGS full (Ours)}     & 6.88  & 27.5 s  & 5.5 s   & 1.10 s \\
\textbf{PD-4DGS Layer~0 only}    & \textbf{0.436} & \textbf{1.74 s} & \textbf{0.35 s} & \textbf{0.07 s} \\
\bottomrule
\end{tabular}
\end{table}

At 2~Mbps, monolithic models take tens of seconds to over ten minutes for the first frame (MoDec-GS 73.5~s, SC-GS 929.6~s), far exceeding the 3--5~s interactive window. PD-4DGS downloads the full model in just 27.5~s (a quarter of MoDec-GS); loading only the $\sim$0.44~MB base layer first delivers a renderable static snapshot in $T_{\text{first}} \approx 1.7$~s---the first time interactive deployment of 4DGS is practical at mobile bandwidths. On broadband links ($\geq$10~Mbps), every incremental layer arrives in $<$1~s, making progressive quality refinement user-imperceptible.

\paragraph{Layer-by-layer refinement and robustness to bandwidth fluctuation.}
The monotonic ordering $\text{PSNR}(\ell_0) \le \text{PSNR}(\ell_1) \le \text{PSNR}(\ell_2)$ verified in Table~\ref{tab:abl-mono} means that receiving Layer~0 yields a static snapshot, $+$Layer~1 a coarsely animated scene, and $+$Layer~2 the full-fidelity dynamic experience---\emph{from a single training run}, making PD-4DGS natively compatible with mainstream ABR protocols unlike single-rate 3DGS compression~\citep{fan2024lightgaussian,niedermayr2024compressed,morgenstern2023compact}. The on-demand nature of HDD is particularly valuable under bandwidth fluctuation: if throughput collapses mid-download the client ``freezes'' at the received layer and keeps rendering; once bandwidth recovers only the next incremental layer is fetched to seamlessly upgrade the picture. This gives 4DGS the same streaming robustness that progressive 3DGS compression~\citep{chen2024hac,disario2025gode,shi2024lapisgs} enjoys on static content, filling a long-standing gap in dynamic-scene streaming.

\section{Limitations and Broader Impacts}

HDD's fixed three-layer topology is most effective on scenes that contain both global and local motion; near-rigid scenes (e.g., the Dycheck teddy) gain only marginally from Layer~2, and the client can \emph{truncate} levels but cannot \emph{insert} new ones at runtime. Following established practice in 3DGS~\citep{kerbl2023threed}, Deformable 3DGS, 4DGS, SC-GS, and MoDec-GS, every number reported here is a single training run per scene (1.5--2~h each); a multi-seed study is left as future work. Our evaluation is limited to the Dycheck iPhone benchmark; extension to other monocular datasets (e.g., HyperNeRF~\citep{park2021hypernerf}), multi-view captures and unbounded outdoor scenes---where Layer~1 may take on more parallax-compensation responsibility---requires further investigation. As for broader impacts, the 60\% streamed-volume reduction and hundred-second$\to$second latency drop enable immersive mobile VR/AR, remote 3D collaboration, and equitable access to volumetric content in low-bandwidth regions, while cutting the energy and network footprint of 4D streaming; the method is non-generative and does not lower the fabricated-content barrier, but the efficient pipeline could in principle be misused for surveillance or unauthorised volumetric capture at scale, so users must comply with local portraiture-rights and data-governance laws. We will release the source code under a non-commercial research licence.

\section{Conclusion}

PD-4DGS is the first framework that brings progressive compression and on-demand transmission to 4DGS. Hierarchical Deformation Decomposition (HDD) restructures the rendering pipeline into three independently transmittable layers; attribute-level R-DO and temporal mask consistency jointly compress anchor attributes and remove low-bitrate flicker; and capacity-weighted rollout training driven by a learnt $\rho$ resolves deformation-network under-training and monotonicity collapse \emph{without any per-scene hyperparameter}. On the Dycheck iPhone benchmark PD-4DGS cuts the streamed bitstream by 62.6\% relative to the strongest dynamic baseline at matched fidelity and delivers a renderable static first frame in $\sim$1.7~s on a 2~Mbps link, making 4DGS natively compatible with adaptive-bitrate streaming for the first time.

\renewcommand{\bibfont}{\small}   
\setlength{\bibsep}{2pt plus 0.3ex}
\bibliographystyle{unsrtnat}
\bibliography{refs}


\end{document}